\documentclass[10pt,journal,compsoc]{IEEEtran}

\ifCLASSOPTIONcompsoc
  \usepackage[nocompress]{cite}
\else
  \usepackage{cite}
\fi

\usepackage{amsmath}
\usepackage{algorithmic}
\usepackage{array}
\usepackage{url}
\usepackage[pdftex]{graphicx}
\usepackage{hyperref}
\usepackage{booktabs}
\usepackage{comment}
\usepackage{wrapfig}
\usepackage{multirow}
\usepackage{listings}
\usepackage{enumitem}

\begin{document}

\title{AutoGL: A Library for Automated Graph Learning}
\author{Ziwei~Zhang\IEEEauthorrefmark{1},~\IEEEmembership{Member,~IEEE,}
        Yijian~Qin\IEEEauthorrefmark{1},
        Zeyang~Zhang\IEEEauthorrefmark{1},
        Chaoyu~Guan\IEEEauthorrefmark{1},
        Jie~Cai\IEEEauthorrefmark{2},
        Heng~Chang\IEEEauthorrefmark{2},
        Jiyan~Jiang\IEEEauthorrefmark{2},
        Haoyang~Li\IEEEauthorrefmark{2},
        Zixin~Sun\IEEEauthorrefmark{2},
        Beini~Xie\IEEEauthorrefmark{2},
        Yang~Yao\IEEEauthorrefmark{2},
        Yipeng~Zhang\IEEEauthorrefmark{2},
        Xin~Wang\IEEEauthorrefmark{3},~\IEEEmembership{Member,~IEEE,}
        and~Wenwu~Zhu\IEEEauthorrefmark{3},~\IEEEmembership{Fellow,~IEEE}
\IEEEcompsocitemizethanks{\IEEEcompsocthanksitem All authors are with the Department
of Computer Science and Technology, Tsinghua University, Beijing,
China. E-mail: autogl@tsinghua.edu.cn
\protect\\ \IEEEauthorrefmark{1}: these authors contributed equally.
\protect\\ \IEEEauthorrefmark{2}: these authors are listed alphabetically.
\protect\\ \IEEEauthorrefmark{3}: corresponding authors.
}
\thanks{Manuscript received XXX}}

\markboth{Journal of \LaTeX\ Class Files,~Vol.~14, No.~8, August~2015}%
{Shell \MakeLowercase{\textit{et al.}}: Bare Demo of IEEEtran.cls for Computer Society Journals}

\IEEEtitleabstractindextext{
\begin{abstract}
Recent years have witnessed an upsurge in research interests and applications of machine learning on graphs. However, manually designing the optimal machine learning algorithms for different graph datasets and tasks is inflexible, labor-intensive, and requires expert knowledge, limiting its adaptivity and applicability. Automated machine learning (AutoML) on graphs, aiming to automatically design the optimal machine learning algorithm for a given graph dataset and task, has received considerable attention. However, none of the existing libraries can fully support AutoML on graphs. To fill this gap, we present Automated Graph Learning (AutoGL), the first dedicated library for automated machine learning on graphs. AutoGL is \textbf{open-source}, \textbf{easy to use}, and \textbf{flexible to be extended}. Specifically, we propose a three-layer architecture, consisting of backends to interface with devices, a complete automated graph learning pipeline, and supported graph applications. The automated machine learning pipeline further contains five functional modules: auto feature engineering, neural architecture search, hyper-parameter optimization, model training, and auto ensemble, covering the majority of existing AutoML methods on graphs. For each module, we provide numerous state-of-the-art methods and flexible base classes and APIs, which allow easy usage and customization. We further provide experimental results to showcase the usage of our AutoGL library. 
We also present AutoGL-light, a lightweight version of AutoGL to facilitate customizing pipelines and enriching applications, as well as benchmarks for graph neural architecture search. The codes of AutoGL are publicly available at \url{https://github.com/THUMNLab/AutoGL}.
\end{abstract}

}

\maketitle
\IEEEdisplaynontitleabstractindextext
\IEEEpeerreviewmaketitle

\IEEEraisesectionheading{\section{Introduction}\label{sec:introduction}}
\IEEEPARstart{M}achine learning on graphs has drawn increasing popularity in the last decade~\cite{zhang2020deep}. For example, graph neural networks (GNNs)~\cite{zhou2018graph,wu2020comprehensive} are the de facto standards and have shown successes in a wide range of applications such as recommendation~\cite{ma2019, li2021intention}, community detection~\cite{wang2017community}, traffic forecasting~\cite{jiang2021graph}, physical simulations~\cite{shlomi2020graph}, geometric data analysis~\cite{gnn}, bioinformatics~\cite{su2020network}, and combinatorial optimization~\cite{bengio2020machine}. 

However, as the literature booms and graph tasks become ever more diverse, it becomes increasingly difficult to manually design the optimal machine learning algorithm for a given graph task. Therefore, there is an urgent need and recent research interest in automated machine learning (AutoML) on graphs~\cite{zhang2021automated}. Essentially, AutoML on graphs combines the strengths of graph-based machine learning and AutoML techniques~\cite{yao2018taking} to automate the design of graph-based machine learning algorithms. Considerable successes have been shown in hyper-parameter optimization (HPO)~\cite{autone} and neural architecture search (NAS)~\cite{graphnas} for graph learning algorithms~\cite{gao2023hgnas++,gao2022graphnas++,wang2023automated}.

Public libraries are critical to facilitating and advancing the research and applications of AutoML on graphs. Several libraries and toolkits exist for machine learning on graphs, such as PyTorch Geometric~\cite{pyg}, Deep Graph Library~\cite{dgl}, GraphNets~\cite{graphnet}, AliGraph~\cite{aligraph}, and PBG~\cite{pbg}. Besides, AutoML libraries such as AutoKeras~\cite{autokeras}, AutoSklearn~\cite{autosklearn}, Hyperopt~\cite{hyperopt}, and NNI~\cite{nni} are also available. Unfortunately, integrating these libraries is non-trivial due to the challenges of AutoML on graphs (see \cite{zhang2021automated} for a comprehensive survey). Currently, there are no public libraries for AutoML on graphs, to the best of our knowledge. 

\begin{figure*}[t]
\centering
\includegraphics[width=\textwidth]{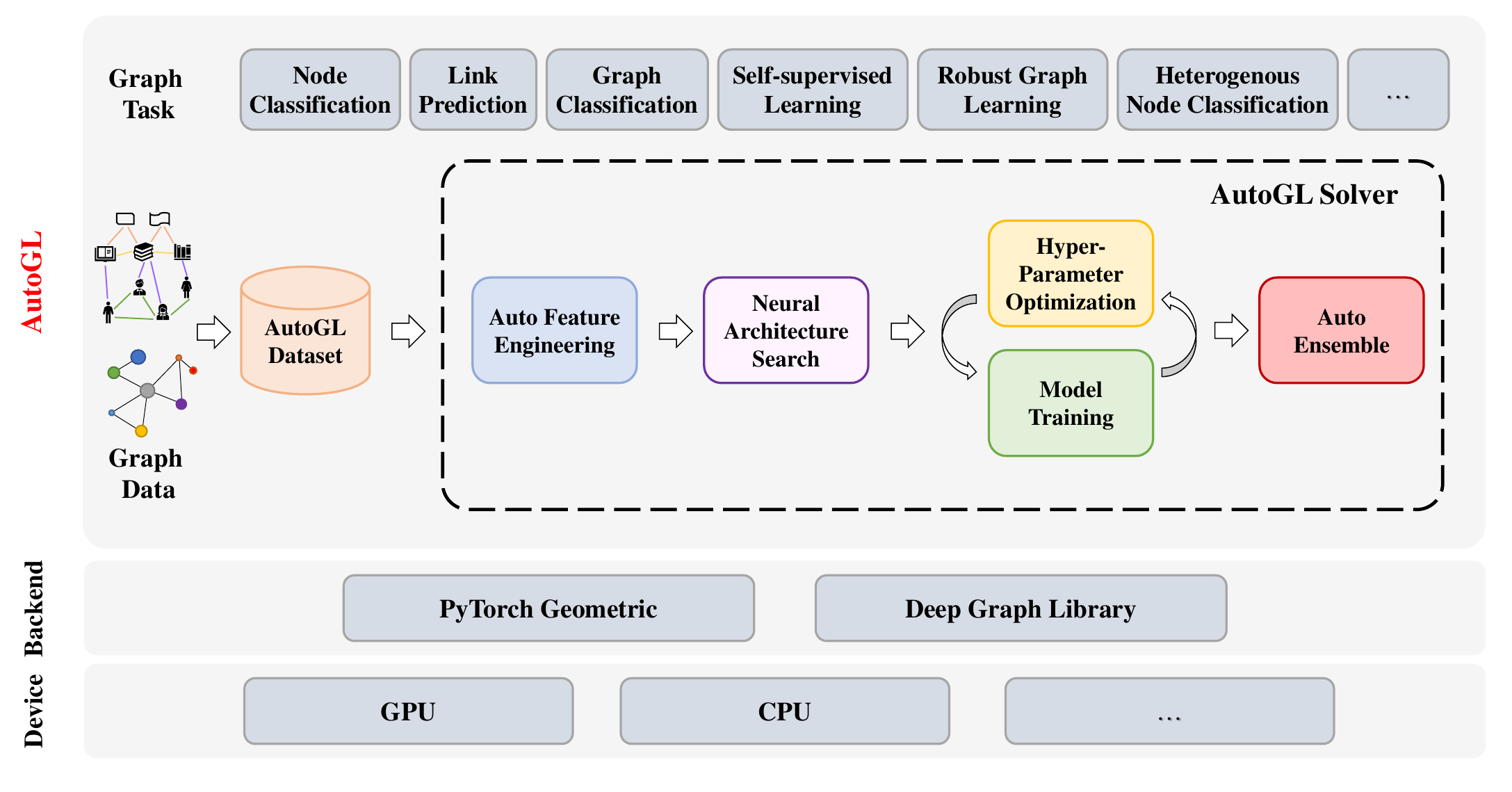}
\caption{An overall framework of AutoGL. The architecture is divided into three layers. At the bottom layer, AutoGL uses the existing graph learning libraries PyTorch Geometric~\cite{pyg} and Deep Graph Learning~\cite{dgl} as backends to interact with the hardware devices. At the middle layer, we design a comprehensive automated graph learning solution. First, we use \texttt{AutoGL Dataset} class to manage diverse graph data, including homogeneous and heterogeneous graphs. Then, we utilize \texttt{AutoGL Solver}, a high-level API to control the workflow of automated graph learning, covering five main functional blocks: \texttt{Auto Feature Engineering}, \texttt{Neural Architecture Search}, \texttt{Hyper-Parameter Optimization}, \texttt{Model Training}, and \texttt{Auto Ensemble}. Our design covers the mainstream automated graph learning methods. At the top layer, we support various graph tasks, including node classification, link prediction, graph classification, self-supervised graph learning, robust graph learning, heterogeneous node classification, etc. }
\label{fig:workflow}
\end{figure*}

To tackle this problem, we present Automated Graph Learning (AutoGL), the first dedicated framework and library for automated machine learning on graphs. The overall framework of AutoGL is shown in Figure~\ref{fig:workflow}. We summarize and abstract AutoML on graphs into a three-layer architecture. On the bottom layer, we use different graph learning libraries as backends (Section~\ref{sec:backend}), supporting two representative graph libraries: PyTorch Geometric (PyG)~\cite{pyg} and Deep Graph Library (DGL)~\cite{dgl}. On the middle layer, we propose a complete automated graph learning pipeline composed of graph datasets (Section~\ref{sec:dataset}) and \texttt{AutoGL Solver} (Section~\ref{sec:solver}), a high-level API to control the automated pipeline. The \texttt{AutoGL solver} further consists of five functional modules: Auto Feature Engineering (Section~\ref{sec:fe}), Neural Architecture Search (Section~\ref{sec:nas}), Hyper-Parameter Optimization (Section~\ref{sec:hpo}), Model Training (Section~\ref{sec:model}), and Auto Ensemble (Section~\ref{sec:ensemble}). For each module, we provide plenty of state-of-the-art algorithms, standardized base classes, and high-level APIs for easy usage and flexible customization. On the top layer (Section~\ref{sec:tasks}), we support various graph tasks, such as node classification, link prediction, graph classification, self-supervised learning, robust graph learning, and heterogeneous node classification. 

In summary, AutoGL has the following characteristics: 
\begin{itemize}
    \item \textbf{Open source:} The code\footnote{\url{https://github.com/THUMNLab/AutoGL/}} and detailed documentation\footnote{\url{https://mn.cs.tsinghua.edu.cn/AutoGL}} of AutoGL, as well as its extension AutoGL-light\footnote{\url{https://github.com/THUMNLab/AutoGL-light/}}, are available online.
    \item \textbf{Easy to use:} AutoGL is designed to be user-friendly. Users can conduct quick AutoGL experiments with less than ten lines of code. We also integrate AutoGL with PyTorch Geometric and Deep Graph Library, two widely-used libraries for graph machine learning.
    \item \textbf{Flexible to be extended:} The modular design, high-level base class APIs, and extensive documentation of AutoGL allow flexible and easily customized extensions.
\end{itemize}

\section{Automated Graph Learning}
In this section, we introduce AutoGL designs in detail. AutoGL is designed in a modular and object-oriented fashion to enable clear logic flows, easy usage, and flexible extensions. All the APIs exposed to users are abstracted in a high-level fashion to avoid redundant re-implementation of models, algorithms, and train/evaluation protocols. All the modules in AutoGL have taken into account the unique characteristics of machine learning on graphs.

\subsection{Backend Supports}\label{sec:backend}
AutoGL uses graph learning libraries as backends and interfaces with the hardware and devices. Currently, AutoGL supports both PyTorch Geometric (PyG)~\cite{pyg} and Deep Graph Library (DGL)~\cite{dgl}, two widely adopted libraries, as the backend to enable users from both communities benefiting from the automation of graph learning. Users can easily check the backend and declare the backend using environment variables. If no backend is specified, AutoGL will use the backend in the environment. If both Deep Graph Library and PyTorch Geometric are installed, the default backend will be Deep Graph Library.

\subsection{AutoGL Dataset} \label{sec:dataset}
Next, we briefly introduce our dataset management. AutoGL Dataset generally adheres to its backend with slight modifications. Specifically, we support most homogeneous graphs from PyG~\cite{pyg} and DGL~\cite{dgl}, and Open Graph Benchmark \cite{ogb},
including widely adopted node classification datasets such as Cora, CiteSeer, PubMed~\cite{sen2008collective}, Amazon Computers, Amazon Photo, Coauthor CS, Coauthor Physics~\cite{amazon}, Reddit~\cite{graphsage}, and graph classification datasets from TU dataset~\cite{tudataset}, such as MUTAG~\cite{mutag}, PROTEINS~\cite{borgwardt2005protein}, IMDB-B, IMDB-M, COLLAB~\cite{graphkernel}, etc. We also support some heterogeneous datasets from DGL for the heterogeneous node classification task. Table~\ref{tab:dataset} summarizes the statistics of the supported datasets.

\begin{table*}[th]
    \centering
    \caption{The statistics of the supported datasets. For undirected graphs, each edge is counted once. For datasets with more than one graph, \#Nodes and \#Edges are the average numbers of all the graphs. \#Features correspond to node features by default, and edge features are specified. }
    \begin{tabular}{lllllll}
    \toprule
    Dataset          & Task                & \#Graphs & \#Nodes & \#Edges & \#Features & \#Classes \\ \midrule
    Cora             & Node  & 1  & 2,708    & 5,278    & 1,433 & 7   \\  
    CiteSeer         & Node  & 1  & 3,327    & 4,552    & 3,703 & 6  \\   
    PubMed           & Node  & 1  & 19,717   & 44,324   & 500   & 3  \\   
    Reddit           & Node  & 1  & 232,965  &114,615,892& 602   & 41 \\  
    Amazon Computers & Node  & 1  & 13,752   & 245,861  & 767   & 10 \\   
    Amazon Photo     & Node  & 1  & 7,650    & 119,081  & 745   & 8  \\  
    Coauthor CS      & Node  & 1  & 18,333   & 81,894   & 6,805 & 15 \\   
    Coauthor Physics & Node  & 1  & 34,493   & 247,962  & 8,415 & 5  \\   
    OGBN-arxiv       & Node  & 1  & 169,343    & 1,166,243     & 128     & 40 \\ 
    OGBN-proteins    & Node  & 1  & 132,534    & 39,561,252    & 8(edge) & 112\\ 
    OGBN-products    & Node  & 1  & 2,449,029  & 61,859,140    & 100     & 47 \\
    OGBN-papers100M  & Node  & 1  & 111,059,956& 1,615,685,872 & 128     & 172\\ 
    OGBL-ddi         & Link  & 1  & 4,267	   & 1,334,889	   & -    & - \\ 
    OGBL-collab      & Link  & 1  & 235,868	   & 1,285,465	   & 128  & - \\
    OGBL-ppa         & Link  & 1  & 576,289	   & 30,326,273	   & 58   & - \\
    OGBL-citation2   & Link  & 1  & 2,927,963  & 30,561,187    & 128  & - \\
    Mutag            & Graph & 188     & 17.9  & 19.8    & 7, 4(edge) & 2 \\ 
    PTC              & Graph & 344     & 14.3  & 14.7    & 18, 4(edge) & 2 \\ 
    ENZYMES          & Graph & 600     & 32.6  & 62.1    & 3 & 6 \\ 
    PROTEINS         & Graph & 1,113   & 39.1  & 72.8    & 3 & 2 \\ 
    NCI1             & Graph & 4,110   & 29.8  & 32.3    & 37 & 2 \\ 
    COLLAB           & Graph & 5,000   & 74.5  & 2,457.8 & - & 3 \\ 
    IMDB-B           & Graph & 1,000   & 19.8  & 96.5    & - & 2 \\ 
    IMDB-M           & Graph & 1,500   & 13.0  & 65.9    & - & 3 \\ 
    REDDIT-B         & Graph & 2,000   & 429.6 & 497.8   & - & 2 \\ 
    REDDIT-MULTI-5K   & Graph & 4,999   & 508.5 & 594.9   & - & 5 \\ 
    REDDIT-MULTI-12K  & Graph & 11,929  & 391.4 & 456.9   & - & 11\\ 
    OGBG-molhiv      & Graph & 41,127  & 25.5  & 27.5    & 9, 3(edge) & 2 \\ 
    OGBG-molpcba     & Graph & 437,929 & 26.0  & 28.1    & 9, 3(edge) & 128\\ 
    OGBG-ppa         & Graph & 158,100 & 243.4 & 2,266.1 & 7(edge) & 37 \\ 
    \bottomrule
    \end{tabular}
    \label{tab:dataset}
\end{table*}
Users can also easily customize datasets following our documentation. We expect our dataset design to support smooth transitions from conducting machine learning on graphs to AutoML on graphs.

\subsection{AutoGL Solver}\label{sec:solver} 
On top of the modules mentioned above, we provide another high-level API Solver to control the overall pipeline. In Solver, all the modules of AutoGL are integrated systematically to form the final model. Solver receives the configurations of functional modules, including a feature engineering module, a model list, a NAS module, an HPO module, and an ensemble module as initialization arguments to build an Auto Graph Learning pipeline. Given a dataset and a task, Solver first performs auto feature engineering to clean and augment the input data, then uses NAS to search for the optimal model and optimize all the given models using the model training and HPO module. At last, the optimized best models will be combined by the Auto Ensemble module to form the final model. Currently, we mainly support three Solvers for three kinds of graph tasks: node classifier, link predictor, and graph classifier. Two specific Solvers, a heterogeneous node classifier and a self-supervised graph classifier are also supported for the specific task. 

Solver also provides global controls of the AutoGL pipeline. For example, the time budget can be explicitly set to restrict the maximum time cost, and the training/evaluation protocols can be selected from plain dataset splits or cross-validation.

\subsection{Auto Feature Engineering}\label{sec:fe}
The graph data is first processed by the auto feature engineering module, where various nodes, edges, and graph-level features can be automatically added, compressed, or deleted to help boost the graph learning process afterward. Graph topological features can also be extracted to utilize graph structures better.

Currently, we support 26 feature engineering operations abstracted into four categories: generators, selectors, graph features, and structure editors. The generators aim to create new node and edge features based on the current node features and graph structures. The selectors automatically filter out and compress features to ensure they are compact and informative. Graph features focus on generating graph-level features. The structure editors aim to modify the graph structure, e.g., adding or deleting edges.

\begin{table}[t]
\caption{Supported generators in the auto feature engineering module.}\label{tab:generators}
\centering
\begin{tabular}{l|l}
\toprule
    Name & Description \\ \midrule
    \texttt{graphlet} & Local graphlet numbers\\ 
    \texttt{eigen}    & EigenGNN features. \\ 
    \texttt{pagerank} & PageRank scores. \\ 
    \texttt{PYGLocalDegreeProfile} & Local Degree Profile features \\ 
    \texttt{PYGNormalizeFeatures} & Row-normalize all node features \\ 
    \texttt{PYGOneHotDegree} & One-hot encoding of node degrees. \\ 
    \texttt{onehot} & One-hot encoding of node IDs \\ \bottomrule
\end{tabular}
\end{table}

\begin{table}[t]
\centering
\caption{Supported selectors in the auto feature engineering module.}\label{tab:selectors}
\begin{tabular}{l|p{5cm}}
\hline
    Name & Description \\ \hline
    \texttt{SeFilterConstant} & Delete constant features and node features that appear only once \\ \hline
    \texttt{gbdt} & Select the top-k important node features ranked by Gradient Boosting Decision Tree \\ \hline
\end{tabular}
\end{table}
We summarize the supported generators in Table~\ref{tab:generators}, including Graphlets~\cite{milo2002network}, EigenGNN~\cite{zhang2020eigen}, PageRank~\cite{pagerank}, local degree profile, normalization, one-hot degrees, and one-hot node IDs. For selectors, GBDT~\cite{NIPS2017_6449f44a} and FilterConstant are supported. An automated feature engineering method DeepGL~\cite{rossi2018deep} is also supported, functioning as both a generator and a selector. For graph features, Netlsd~\cite{tsitsulin2018netlsd} and a set of graph feature extractors implemented in NetworkX~\cite{networkx} are wrapped, e.g., \texttt{NxTransitivity},  \texttt{NxAverageClustering}, etc. For structure editors, we support GCNSVD~\cite{entezari2020all} and GCNJaccard~\cite{wu2019adversarial}.

We also provide convenient wrappers that support feature engineering operations in PyTorch Geometric~\cite{pyg} and NetworkX~\cite{networkx}. Users can easily customize feature engineering methods by inheriting from the class \texttt{BaseGenerator}, \texttt{BaseSelector}, and \texttt{BaseGraph}, or \texttt{BaseFeatureEngineer} if the methods do not fit in our categorization.

\subsection{Neural Architecture Search}\label{sec:nas}
NAS, aiming to automatically design the optimal neural architecture for a given task and dataset, has been the center of AutoML research in the deep learning era. Similarly, graph NAS has also aroused considerable attention in the past few years. AutoGL has supported various graph NAS algorithms. Technically, NAS is usually divided into three key modules~\cite{elsken2019neural}: search space, search strategy, and performance estimation strategy. Therefore, we also divide the implementation of AutoGL for graph NAS into three parts: algorithm, space, and estimator, corresponding to the three modules. respectively. Since these three modules are relatively modularized, different choices in different modules can be composed under certain constraints. If users want to implement their own NAS process, they can change any of those parts according to the demand. Next, we elaborate on the designs of AutoGL in these three aspects.  

\textbf{The search space} describes all possible architectures to be searched. A larger search space may contain more effective architecture but also demands more effort to explore. Therefore, an ideal search space should balance effectiveness and efficiency. Besides, human knowledge can be utilized to design a more effective search space and alleviate the burdens of search strategies.

For graph NAS, there are mainly two parts of the space~\cite{zhang2021automated}: the connectivity patterns between the input and output (also called the macro search space) and the message-passing operations (also called the micro search space). In AutoGL, the space definition is based on a mutable fashion used in NNI~\cite{nni}, which is defined as a model inheriting BaseSpace There are mainly two ways to define the search space: one can be searched with the one-shot fashion using a super network and the other cannot. AutoGL has supported several representative search spaces, including the classical sequential GNNs with various message-passing operations~\cite{zhang2020deep}, the full and macro search space of GraphNAS~\cite{graphnas}, AutoAttend~\cite{guan2021autoattend}, etc. Users can also easily define their own search space. If the search space is compatible with one-shot search strategies, users need to support constructing the super network. Otherwise, if the search space does not need to support the one-shot fashion, the search space is more like normal models with minor changes, but the supported search strategies are limited.

\textbf{The search strategy} controls how to explore the search space and automatically find the optimal architecture. On the one hand, it is desirable to find well-performing architectures as efficiently as possible, while on the other hand, convergence to a region of suboptimal architectures should be avoided. Therefore, the search strategy encompasses the classical exploration-exploitation trade-off problem. There are three major categories of search strategies in NAS: reinforcement learning (RL), evolution algorithm (EA), and differentiable algorithms~\cite{elsken2019neural}. Recently, differentiable algorithms are more preferred since different operations can be jointly optimized in the super network, leading to improved search efficiency. Currently, AutoGL supports both general-purpose NAS search strategies, such as Random~\cite{li2020random}, reinforcement learning (RL)~\cite{zoph2016neural}, evolutionary algorithm (EA)~\cite{liu2021survey}, differentiable architecture search (DARTS)~\cite{liu2018darts}, efficient NAS (ENAS)~\cite{pham2018efficient}, single-pass one-shot NAS (SPOS)~\cite{guo2020single}, Hardware-aware NAS~\cite{benmeziane2021hardware}, and AutoAttend~\cite{guan2021autoattend}, as well as strategies specifically designed for graph NAS, such as GraphNAS~\cite{graphnas}, AutoGNN~\cite{zhou2022auto}, structure learning graph NAS GASSO~\cite{qin2021graph} and robust graph NAS G-RNA~\cite{xie2023adversarially}.

\textbf{The performance estimation strategy} gives the performance of architectures when it is explored. Since there are lots of architectures that need to be estimated in the whole search process, the estimation strategy is desired to be efficient to save computational resources while being accurate. The simplest option is to perform standard training and testing of the architectures on the input data. Though this training-from-scratch method can give the most accurate performance estimation, its computational usage and efficiency are severely limited. Another widely adopted strategy is to use the super network as the performance estimator. Since all possible architectures are fused into a large network, this method is also known as the one-shot performance estimation. Users can also implement their own performance estimation methods. 

\subsection{Hyper-Parameter Optimization}\label{sec:hpo}
The HPO module aims to automatically search for the best hyper-parameters of a specified model and training process, including but not limited to model hyper-parameters such as the number of layers, the dimensionality of node representations, the dropout rate, the activation function, and training hyper-parameters such as the optimizer, the learning rate, the weight decay, and the number of epochs. The hyper-parameters, their types (e.g., integer, numerical, or categorical), and feasible ranges can be easily set.

We have supported various HPO algorithms, including algorithms specified for graph data like AutoNE~\cite{autone} and general-purpose algorithms from advisor library~\cite{golovin2017google}, including grid search, random search~\cite{randomsearch}, Anneal algorithm~\cite{otten2012annealing}, Bayes Optimization~\cite{wu2019hyperparameter}, Tree Parzen Estimator (TPE)~\cite{tpe}, Covariance Matrix Adaptation Evolution Strategy (CMA-ES)~\cite{arnold2010active}, Multi-Objective CMA-ES (MO-CMA-ES)~\cite{voss2010improved}, Quasi random~\cite{bratley1994programs}, etc. Users can customize HPO algorithms by inheriting from the \texttt{BaseHPOptimizer} class.

\subsection{Model Training}\label{sec:model}
This module handles the training and evaluation process of graph machine learning tasks with two functional sub-modules: Model and Trainer. The Model handles the construction of graph machine learning models, e.g., GNNs, by defining learnable parameters and the forward pass. The Trainer controls the optimization process for the given model. Common optimization methods are packaged as high-level APIs to provide neat and clean interfaces. More advanced training controls and regularization methods in graph tasks like early stopping and weight decay are also supported.

The model training module supports both node-level, edge-level, and graph-level tasks, e.g., node classification, link prediction, and graph classification. Commonly used models for node classification such as Graph Convolutional Network (GCN)~\cite{gcn}, Graph Attention Network (GAT)~\cite{gat}, and Graph Sampling and Aggregation (GraphSAGE)~\cite{graphsage}, Graph Isomorphism Network (GIN)~\cite{gin}, and pooling methods such as Top-K Pooling~\cite{topk} are supported. Besides, we have also supported a decoupled design by abstracting and unifying the graph learning model into an encoder and a decoder~\cite{hamilton2017representation} so that methods can be used for different tasks, e.g., using GCN, GAT, and GraphSAGE as the encoder to link prediction and graph classification task. AutoGL also supports heterogeneous trainers for the heterogeneous node classification task and self-supervised learning trainers. 

Users can easily implement their own graph models by inheriting from the \texttt{BaseModel} class and add customized tasks or optimization methods by inheriting from \texttt{BaseTrainer}.

\subsection{Auto Ensemble}\label{sec:ensemble}
This module can automatically integrate the optimized individual models (referred to as base learners) to form a more powerful final model. Currently, we have adopted two kinds of ensemble methods: voting and stacking. 

Voting is a simple yet powerful ensemble method that directly averages the output of individual models using learned weights.  Given an evaluation metric, the weights of base learners are specified to maximize the validation score. Specifically, we first find a collection of base learners with equal weights via a greedy search, then specify the weights in the voter by the number of occurrences in the collection. 

Stacking trains another meta-model to combine the output of models and find an optimal combination of these base learners. We have supported General Linear Models (GLM) and Gradient Boosting Machines (GBM) as meta-models. 

Users can create their own ensemble method by inheriting the 
\texttt{BaseEnsembler} class and overloading \texttt{fit} and \texttt{ensemble} functions.

\subsection{Graph Task}\label{sec:tasks}
\subsubsection{Node Classification}
Node classification, which aims to predict the labels of nodes based on their features and the graph structure, is a widely adopted graph task. In this section, we use node classification as a showcase to introduce the general usage of AutoGL.

Let us assume users want to conduct auto graph learning on the Cora dataset. First, users can easily get the Cora dataset using the dataset module. The dataset will be automatically downloaded and processed. 
\begin{lstlisting}[language=python, basicstyle=\footnotesize]
from autogl.datasets import build_dataset_from_name
cora = build_dataset_from_name('cora')
\end{lstlisting}
After obtaining the dataset, users can build a node classification solver to handle the auto-training process:
\begin{lstlisting}[language=python, basicstyle=\footnotesize]
from autogl.solver import AutoNodeClassifier
solver = AutoNodeClassifier(
  feature_module='deepgl',
  graph_models=['gcn', 'gat'],
  hpo_module='anneal',
  ensemble_module='voting')
\end{lstlisting}
In the above code, we build a node classification solver, which uses deepgl as its feature engineer and uses anneal hyperparameter optimizer to optimize the given two models, GCN and GAT. The derived models will then be ensembled using the voting method. Note that we have omitted the NAS module in this example. Then, users can fit the solver and then check the leaderboard:
\begin{lstlisting}[language=python, basicstyle=\footnotesize]
solver.fit(cora, time_limit=3600)
solver.get_leaderboard().show()    
\end{lstlisting}
Specifically, the time limit is set to 3600 seconds so that the whole auto graph learning process will not exceed 1 hour. \texttt{solver. show()} will present the models maintained by the solver, with their performances on the validation dataset. Lastly, users can make the predictions and evaluate the results using the evaluation functions provided:
\begin{lstlisting}[language=python, basicstyle=\footnotesize]
from autogl.module.train import Acc
predicted = solver.predict_proba()
print('Test acc:', Acc.evaluate(predicted,
     Cora.data.y[cora.data.test_mask].cpu().numpy()))
\end{lstlisting}
So far, we have finished designing an automated graph learning method for the node classification task and obtained the final results.

\subsubsection{Link Prediction}
Link prediction, aiming to predict the missing links in a graph, is a typical edge-level graph task with important real-world applications, such as recommendation systems. In AutoGL, we use a Solver to control the overall pipeline for link prediction. Most design principles of the link prediction Solver are similar to the node classification Solver, except that the training and evaluation are calculated on pairs of nodes. As explained in Section~\ref{sec:model}, our link prediction Solver also supports decoupling the model into an encoder and a decoder, where the encoder is responsible for learning low-dimensional vector representations of nodes and the decoder further calculates the probability of two nodes forming an edge based on the representation of these nodes.

\subsubsection{Graph Classification}
Instead of focusing on individual or pairs of nodes, the goal of graph classification is to predict the label of the whole graph. In AutoGL, another Solver is designed to support the graph classification task. As the information of nodes needs to be aggregated into a graph-level representation, the Solver needs to cover the design of pooling functions, also known as the readout function~\cite{zhang2020deep}. Currently, AutoGL mainly supports two pooling functions: basic mathematical operations such as element-wise maximum, mean, and summation, and top-$k$ pooling~\cite{zhang2018end}.

\subsubsection{Self-supervised Graph Learning}
Self-supervised graph learning~\cite{liu2022graph} is a trending direction for graph machine learning where the model is expected to learn useful knowledge without using any label information. In AutoGL, we have also supported self-supervised graph learning. Note that currently, we only support automated graph machine learning under the self-supervised setting, while how to automate the self-supervised learning per se, e.g., automatically combining multiple self-supervised tasks~\cite{jin2022automated}, is left as future works.  

AutoGL uses a new solver and trainer to implement the graph self-supervised methods, considering that the training and evaluation protocols differ from classical supervised training. Currently, we support Graph Contrastive Learning (GraphCL)~\cite{you2020graph} for the semi-supervised node classification task as a showcase, while more methods can be easily implemented. The key idea of GraphCL is to randomly perturb subgraphs centered at each node and learn node representations using a GNN encoder. Two representations of the same node are considered positive samples, while the representations of different nodes are considered negative samples. Then, a contrastive loss function to distinguish positive and negative samples is optimized to learn the parameters of the GNN encoder. Finally, the GNN encoder and node representations, which are learned in a self-supervised manner, can be applied to downstream graph tasks such as node classification. 

\subsubsection{Robust Graph Learning}
Graph robustness is an important research direction in the field of graph machine learning in recent years~\cite{sun2022adversarial}, and we have integrated robustness-related graph algorithms in AutoGL, which can be easily used in conjunction with other modules. In AutoGL, we divide the methods for graph robustness into three categories, which are placed in different modules for implementation, detailed as follows. 

\textbf{Robust Graph Feature Engineering} aims to generate robust graph features in the data pre-processing phase to enhance the robustness of models. Specifically, we provide two representative structure engineering methods from DeepRobust~\cite{li2021deeprobust} to enhance the robustness of graph machine learning: GCN-SVD~\cite{entezari2020all}, which assumes that the robust graph structure is inherently low-rank and adopts Singular Value Decomposition (SVD) to recover the robust graph structure, and GCN-Jaccard~\cite{wu2019adversarial}, which is based on the homophily assumption and removes edges that connect nodes with small Jaccard similarity of features. 

\textbf{Robust GNNs} are designed at the model level or the training process to enhance the robustness. As a showcase, we incorporate GNNGuard~\cite{zhang2020gnnguard}, which assigns weights to edges based on their usefulness, i.e., edges connecting similar nodes are assigned larger weights, and edges connecting unrelated nodes have smaller weights. Other robust GNN models can be added following similar principles.

\textbf{Robust graph NAS} aims to search for a robust GNN architecture. In AutoGL, we implement Robust Graph Neural Architecture Search (G-RNA)~\cite{xie2023adversarially}, the only robust graph NAS to date. Specifically, G-RNA designs a robust search space for the message-passing mechanism by adding adjacency mask operations, which are inspired by the existing defensive operators. Furthermore, G-RNA defines a robustness metric to guide the search procedure, which helps to filter robust architectures. The details of these two components are as follows.

For the adjacency mask operations, G-RNA includes five mask operations in the search space:
\begin{itemize}[leftmargin=0.5cm]
    \item Identity: keep the same adjacency matrix as in the previous layer, i.e., without adding robustness designs.
    \item Low-Rank Approximation (LRA): reconstruct the adjacency matrix from the top-$k$ components of singular value decomposition, as in GCN-SVD~\cite{entezari2020all}.
    \item Node Feature Similarity (NFS): delete edges that have small Jaccard similarities among node features, as in GCN-Jaccard~\cite{wu2019adversarial}.
    \item Neighbor Importance Estimation (NIE): update mask values with a pruning strategy based on quantifying the relevance among nodes, as in GNNGuard~\cite{zhang2020gnnguard}.
    \item Variable Power Operator (VPO): form a variable power graph from the original adjacency matrix weighted by the parameters of influence strengths as in~\cite{jin2021power}.
\end{itemize}
To measure the robustness of GNN architecture, G-RNA uses the Kullback-Leibler (KL) divergence to measure the prediction difference between clean and perturbed data. A larger robustness score indicates a smaller distance between the prediction of clean data and the perturbed data, and consequently, more robust GNN architectures.

By implementing the search space and search algorithm, AutoGL can support G-RNA and users can also try and test their own robust models.

\subsubsection{Heterogeneous Node Classification}
Heterogeneous graphs, where nodes and edges have different types, are common in the real world, such as recommendation systems, academic graphs, knowledge graphs, etc. AutoGL provides another Solver API, termed \texttt{AutoHeteroNodeClassifier}, for heterogeneous node classification, which is the most common graph task for heterogeneous graphs. Specifically, we use the heterogeneous graph datasets from DGL and encapsulate the training process in the solver, which supports automatic hyper-parameter optimization as well as feature engineering and ensemble. In the provided tutorial, we showcase how to use two representative heterogeneous graph neural network models, Heterogeneous Attention Network (HAN)~\cite{wang2019heterogeneous} and Heterogeneous Graph Transformer (HGT)~\cite{hu2020heterogeneous}, for ACM dataset, where the task is to predict the publishing venue of a paper using the ACM academic graph dataset. Users can implement their own models and use other datasets following the instructions of the tutorial.

\section{Evaluation}
\begin{table}[t]
\centering
\caption{The results of node classification in AutoGL.}\label{exp:semi}
\begin{tabular}{cccc} \toprule
Model & Cora & CiteSeer & PubMed \\ \midrule
GCN & $80.9 \pm 0.7$ & $70.9 \pm 0.7$ & $78.7 \pm 0.6$ \\
GAT & $82.3 \pm 0.7$ & $71.9 \pm 0.6$ & $77.9 \pm 0.4$ \\
GraphSAGE & $74.5 \pm 1.8$ & $67.2 \pm 0.9$ & $76.8 \pm 0.6$ \\
AutoGL & $\mathbf{83.2 \pm 0.6}$ & $\mathbf{72.4 \pm 0.6}$ & $\mathbf{79.3 \pm 0.4}$ \\ \bottomrule
\end{tabular}
\end{table}

\begin{table}[t]
\centering
\caption{The results of graph classification in AutoGL.}\label{exp:graph}
\begin{tabular}{cccc} \toprule
Model & MUTAG & PROTEINS & IMDB-B \\ \midrule
Top-K Pooling & $80.8 \pm 7.1$ & $69.5 \pm 4.4$ & $71.0 \pm 5.5$ \\
GIN & $82.7 \pm 6.9 $ & $66.5 \pm 3.9$ & $69.1 \pm 3.7$ \\
AutoGL & $\mathbf{87.6 \pm 6.0}$ & $\mathbf{73.3 \pm 4.4}$ & $\mathbf{72.1 \pm 5.0}$ \\ \bottomrule
\end{tabular}
\end{table}

\begin{table}[t]
\centering
\caption{The results of graph NAS in AutoGL under the fully supervised setting in AutoGL.}\label{exp:NAS}
\begin{tabular}{cccc} \toprule
Model & Cora & CiteSeer & PubMed \\ \midrule
Random Search & $ 86.47 \pm 0.68$ & $ 78.77 \pm 0.61$ & $87.83 \pm 0.40$ \\
RL & $ 86.37 \pm 0.42$ & $ 77.83 \pm 1.35$ & $ 86.83 \pm 1.50 $ \\
DARTS & $ 86.68 \pm 0.45 $ & $77.84 \pm 0.62$ & $ 88.02 \pm 0.18 $ \\
GASSO & $\mathbf{87.03 \pm  0.18}$ & $\mathbf{79.12 \pm 0.52 }$ & $\mathbf{88.51 \pm 0.34 }$ \\ \bottomrule
\end{tabular}
\end{table}

\begin{table*}[t]
\centering
\caption{The results of different HPO methods for node classification.}\label{exp:hpo:node}
\begin{tabular}{cccccccc} \toprule
\multicolumn{2}{c}{} & \multicolumn{2}{c}{Cora} & \multicolumn{2}{c}{CiteSeer} & \multicolumn{2}{c}{PubMed} \\
Method & Trials & GCN & GAT & GCN & GAT & GCN & GAT \\ \midrule
\multicolumn{2}{c}{None} & $80.9\pm 0.7$ & $82.3\pm0.7$ & $70.9\pm0.7$ & $71.9\pm 0.6$ & $78.7\pm0.6$ & $77.9\pm0.4$ \\ \midrule
\multirow{3}{*}{random} & 1 & $81.0 \pm 0.6$ & $81.4 \pm 1.1$ & $70.4 \pm 0.7$ & $70.1 \pm 1.1$ & $78.3 \pm 0.8$ & $76.9 \pm 0.8$ \\
& 10 & $82.0\pm0.6$ & $82.5\pm0.7$ & $71.5\pm0.6$ & $\bf 72.2\pm0.7$ & $79.1\pm0.3$ & $78.2\pm0.3$ \\
& 50 & $81.8\pm1.1$ & $\bf 83.2\pm0.7$ & $71.1\pm1.0$ & $72.1\pm1.0$ & $\bf 79.2\pm0.4$ & $78.2\pm0.4$ \\ \midrule
\multirow{3}{*}{TPE} & 1 & $81.8 \pm 0.6$ & $81.9 \pm 1.0$ & $70.1 \pm 1.2$ & $71.0 \pm 1.2$ & $78.7 \pm 0.6$ & $77.7 \pm 0.6$ \\
& 10 & $82.0 \pm 0.7$ & $82.3 \pm 1.2$ & $71.2 \pm 0.6$ & $72.1 \pm 0.7$ & $79.0 \pm 0.4$ & $\bf 78.3 \pm 0.4$ \\
& 50 & $\bf 82.1 \pm 1.0$ & $83.2 \pm 0.8$ & $\bf 72.4 \pm 0.6$ & $71.6 \pm 0.8$ & $79.1 \pm 0.6$ & $78.1 \pm 0.4$ \\ \bottomrule
\end{tabular}
\end{table*}

\begin{table*}
\centering
\caption{The results of different HPO methods for graph classification.}\label{exp:hpo:graph}
\begin{tabular}{ccccccc} \toprule
\multirow{2}{*}{HPO} & \multicolumn{2}{c}{MUTAG} & \multicolumn{2}{c}{PROTEINS} & \multicolumn{2}{c}{IMDB-B} \\
 & Top-K Pooling & GIN & Top-K Pooling & GIN & Top-K Pooling & GIN \\ \midrule
None & $76.3\pm7.5$ & $82.7\pm6.9$ & $69.5\pm4.4$ & $66.5\pm3.9$ & $71.0\pm5.5$ & $69.1\pm3.7$\\
random & $82.7\pm 6.8$ & $\bf 87.6 \pm 6.0$ & $\bf 73.3\pm4.4$ & $\bf 71.0\pm5.9$ & $71.5\pm 4.1$ & $\bf 71.3\pm 4.0$ \\
TPE & $\mathbf{ 83.9\pm 10.1}$ & $86.7 \pm 6.2$ & $72.3\pm5.5$ & $71.0\pm7.2$ & $\bf 71.6\pm 2.5$ & $70.2\pm 3.7$ \\ \bottomrule
\end{tabular}
\end{table*}

\begin{table}
\caption{The performance of the ensemble module of AutoGL for the node classification task.}\label{exp:tab:results_ensemble}
\centering
  \begin{tabular}{lcccccc} \toprule
  Base Model & Cora & CiteSeer & PubMed \\ \midrule
  GCN & $81.1 \pm 0.9$ & $69.6\pm 1.1$ & $\bf 78.5 \pm 0.4 $ \\
  GAT & $82.0 \pm 0.5$ & $70.4\pm 0.6$ & $77.7 \pm 0.5 $\\ \midrule
  Ensemble  & $\bf 82.2 \pm 0.4$ & $\bf 70.8 \pm 0.5$ & $\bf 78.5 \pm 0.4$ \\ \bottomrule
  \end{tabular}
\end{table} 
In this section, we provide experimental results to showcase the usage of AutoGL as well as every functional module. Note that we mainly want to showcase the usage of AutoGL and its main functional modules rather than aiming to achieve the new state-of-the-art benchmarks or compare different algorithms. 
For node classification, we use Cora, CiteSeer, and PubMed with the standard dataset splits from \cite{gcn}. For graph classification, we follow the setting in \cite{errica2020fair} and report the average accuracy of 10-fold cross-validation on MUTAG, PROTEINS, and IMDB-B.

\textbf{AutoGL Results} We turn on all the functional modules in AutoGL except NAS, and report the fully automated results in Table~\ref{exp:semi} and Table~\ref{exp:graph}. We use the best single model for graph classification under the cross-validation setting. We observe that in all the benchmark datasets, AutoGL achieves better results than vanilla models, demonstrating the importance of AutoML on graphs and the effectiveness of the proposed pipeline in the released library.

\textbf{Neural Architecture Search} Table~\ref{exp:NAS} reports the results of four representative NAS methods. The results show that GASSO~\cite{qin2021graph}, which is specifically designed for graph data, achieves the best results. Besides, the results obtained in our library are largely aligned with the original results reported in the paper.

\textbf{Hyper-Parameter Optimization} Table~\ref{exp:hpo:node} reports the results of two implemented HPO methods, i.e., random search and TPE~\cite{tpe}, for the semi-supervised node classification task. As shown in the table, as the number of trials increases, both HPO methods tend to achieve better results. Besides, both methods outperform vanilla models without HPO. Note that a larger number of trials do not guarantee better results because of the potential overfitting problem. We further test these HPO methods with ten trials for the graph classification task and report the results in Table~\ref{exp:hpo:graph}. The results generally show improvements over the default hand-picked parameters on all datasets.

\textbf{Auto Ensemble} Table~\ref{exp:tab:results_ensemble} reports the performance of the ensemble module as well as its base learners for the node classification task. We use voting as the example ensemble method and choose GCN and GAT as the base learners. The table shows that the ensemble module achieves better performance than both the base learners, demonstrating the effectiveness of the implemented ensemble module.

\begin{figure*}[t]
\centering
\includegraphics[width=\textwidth]{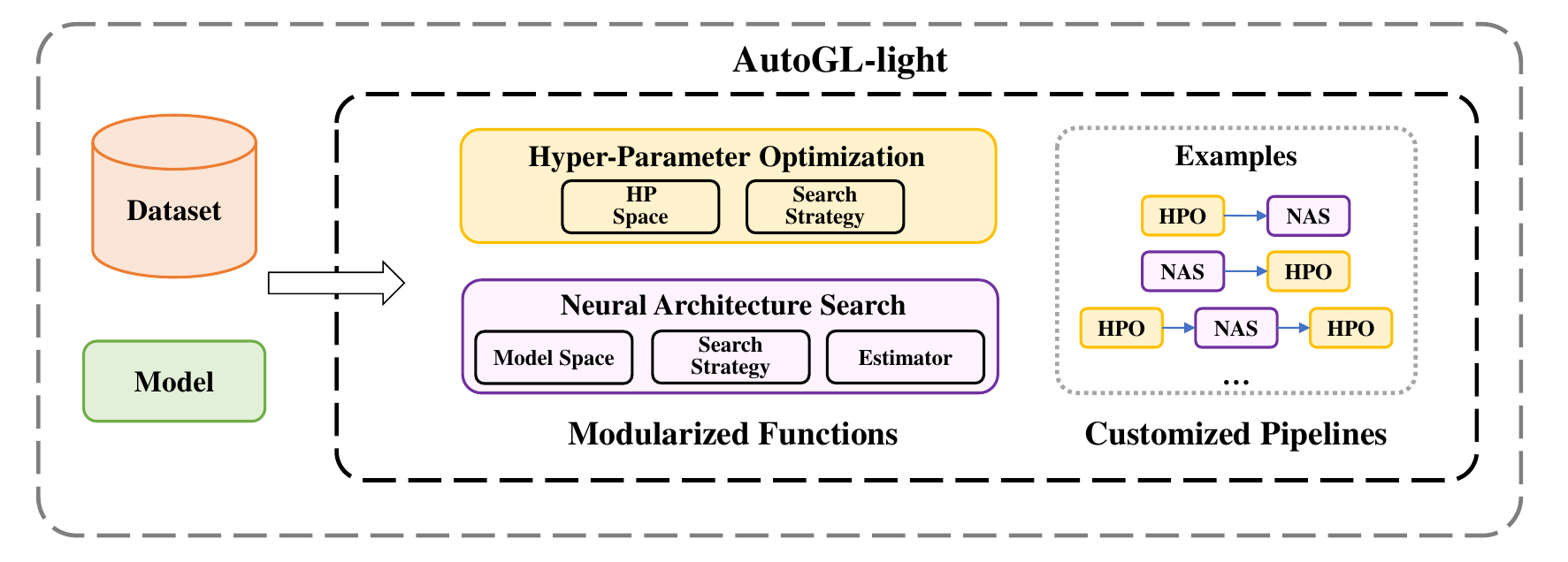}
\caption{An overall framework of AutoGL-light. Compared to AutoGL, AutoGL-light is a more lightweight version with decoupled designs and focuses on two core functionalities of automated graph machine learning: graph HPO and graph NAS. Through modularized functions, AutoGL-light can more flexibly support customized pipelines. We also refactor the code to be more user-friendly to PyG users.  }
\label{fig:lightworkflow}
\end{figure*}

\section{AutoGL-light}
Though AutoGL provides a comprehensive automated pipeline for graph machine learning, the heavy functionality makes its flexibility inevitably limited and the learning cost for users is higher than we anticipated. To further promote the research and usage of automated graph machine learning, we design and develop AutoGL-light\footnote{\url{https://github.com/THUMNLab/AutoGL-light}}, a more lightweight version of AutoGL. The design principle is to decouple the pipelines with modularized functions and focus on more core functionalities while leaving more flexibility to users. An illustration of AutoGL-light is shown in Figure~\ref{fig:lightworkflow}.

Specifically, AutoGL-light includes two main functionalities: graph HPO and graph NAS. Compared to AutoGL, AutoGL-light does not fix the pipeline, i.e., it allows to free incorporation graph HPO and graph NAS at any step of the workflow. We also refactor the code of AutoGL-light to be more user-friendly, especially for users who are already familiar with PyTorch Geometric. We plan to make AutoGL-light more compatible with various graph machine learning libraries in the future. Next, we briefly introduce the designs of graph HPO and graph NAS in AutoGL-light.

Since graph HPO aims to automatically optimize the hyper-parameters of graph machine learning models, as discussed in Section~\ref{sec:hpo}, we divide its designs into two parts: hyper-parameter (HP) space and optimization algorithm. For the HP space, we support three types of search space: linear numerical search space, logarithmic numerical space, and categorical space. The former two are used to optimize numerical hyper-parameters, such as the number of layers, the dimensionality, the learning rate, etc., and the latter is used to optimize categorical hyper-parameters, such as the optimizer. In addition, users can also create their own hyper-parameter space by inheriting the base class. For the optimization algorithms, we support various black-box algorithms as in AutoGL (Section~\ref{sec:hpo}), such as grid search, random search, anneal algorithm, Bayes optimization, CMA-ES, MO-CMA-ES, Quasi-random, TPE, and AutoNE.

As discussed in Section~\ref{sec:nas}, graph NAS is usually composed of three modules: search space, search strategy, and performance estimation. Therefore, we also modularize the NAS function into three parts. Other details are similar to AutoGL, except for code refactoring so that AutoGL-light is no longer dependent on NNI, avoiding potential version incompatibility issues and reducing the learning cost of users. Besides, since the base class of our graph NAS implementation is heavily inspired by NNI, we expect smooth transitions if users are already familiar with NNI.

To promote and showcase the flexibility and easy usage of AutoGL-light, we have implemented several state-of-the-art graph NAS methods, including out-of-distribution generalized graph NAS GRACES~\cite{qin2022graph}, large-scale graph NAS method GAUSS~\cite{guan2022large}, and automated graph Transformer AutoGT~\cite{zhang2023autogt}. Besides, we showcase AutoGL-light for handling various downstream graph tasks by applying AutoGL-light to bio-informatics using graph HPO and graph NAS, including ScGNN~\cite{wang2021scgnn}, MolCLR~\cite{wang2022molecular}, and AutoGNNUQ~\cite{jiang2023uncertainty}. These showcases were supported by GitLink Code Camp\footnote{\url{https://www.gitlink.org.cn/glcc}}.  We aim to incorporate more advanced graph NAS such as~\cite{zhang2024unsupervised,qin2024multi} and more applications such as~\cite{oloulade2023cancer,jin2023dual} in the future.

\section{NAS-Bench-Graph} 

\begin{table*}[t]
\caption{The results of applying two representative graph NAS methods using NAS-Bench-Graph in AutoGL.}\label{exp:benchmark}
\centering
\begin{tabular}{l|cccccccccc}
\toprule
Method & Cora  & CiteSeer & PubMed & CS    & Physics & Photo & Computers & arXiv & proteins \\ \midrule
GNAS   & 82.04$_{0.17}$  & ${70.89}_{ 0.16}$  &  77.79$_{ 0.02}$ & 90.97$_{0.06}$  & 92.43$_{0.04}$ & 92.43$_{0.03}$  & 84.74$_{0.20}$  & 72.00$_{0.02}$  &   {78.71}$_{0.11}$   \\
Auto-GNN  & 81.80$_{0.00}$  & 70.76$_{0.12}$ & 77.69$_{ 0.16}$ &     ${91.04}_{0.04}$  &  92.42$_{0.16}$ &  92.38$_{0.01}$ & 84.53$_{0.14}$ &  {72.13}$_{0.03}$  &  78.54$_{0.30}$   \\ 
 \bottomrule
\end{tabular}
\end{table*}

One challenge for automated graph machine learning research, particularly, graph NAS, is how to fairly and efficiently compare different methods while ensuring reproducibility. NAS-Benchmark contains a series of works tackling this issue by providing the training log, but none of them is designed and suitable for automated graph machine learning. To fill this gap, we have developed NAS-Bench-Graph~\cite{qin2022bench}, the first tabular benchmark for graph NAS. We have also integrated NAS-Bench-Graph with AutoGL, which be elaborated as follows. More details about NAS-Bench-Graph can be referred to our paper~\cite{qin2022bench}.

To use NAS-Bench-Graph, users should define the search space the same as the paper. In NAS-Bench-Graph, we design a search space including 9 macro search space choices and 7 operations including GCN~\cite{gcn}, GAT~\cite{gat}, GraphSAGE~\cite{graphsage}, GIN~\cite{gin}, ChebNet~\cite{defferrard2016convolutional}, ARMA~\cite{bianchi2021graph}, and $k$-GNN~\cite{morris2019weisfeiler}. As a result, the search space contains 26,206 unique architectures (after removing equivalent architectures) and covers many representative GNNs such as GCN, GAT, GraphSAGE, etc., and their variants such as adding residual connections~\cite{xu2018representation} and dense connections~\cite{li2019deepgcns}. The performance of all architectures in the search space has been evaluated on nine representative graph datasets, including Cora, CiteSeer, PubMed~\cite{sen2008collective}, Coauthor-CS, Coauthor-Physics, Amazon-Photo, Amazon-Computers~\cite{amazon}, OGBN-arXiv, and OGBN-Proteins~\cite{ogb}. These datasets cover different sizes from thousands of nodes and edges to hundreds of thousands of nodes and more than 30 million edges, and application domains including citation graphs, co-purchase graphs, and bio-informatics.

Since the detailed metrics of architectures during the training and testing process are recorded in NAS-Bench-Graph, AutoGL can easily support running different NAS search algorithms with NAS-Bench-Graph. Concretely, users can directly get architecture performance estimation from NAS-Bench-Graph instead of training the architectures from scratch. Therefore, users only need to define an estimator get the performance of architectures on the given dataset from the benchmark, and use search algorithms to obtain the optimal architectures. 

Next, we use some sample codes to illustrate the usage of NAS-Bench-Graph in AutoGL:
\begin{lstlisting}[language = python, basicstyle=\footnotesize]
space = BenchSpace().cuda()
space.instantiate(input_dim, output_dim, ops_type)
estimator = BenchEstimator(dataset_name)
alg = GraphNasRL(num_epochs)
model = alg.search(space, dataset=None, estimator)
result = estimator.infer(model._model,None)[0][0]
\end{lstlisting}
In the above codes, we first initialize the search space and performance estimator. Then we choose a NAS search strategy and initialize it. After that, we run the search and inference process. Using NAS-Bench-Graph, graph NAS can be executed with minimum computational resources, and the results are guaranteed to be reproducible. As a showcase, we show the results of applying two graph NAS methods, GraphNAS~\cite{graphnas} and AutoGNN~\cite{zhou2022auto}, using our proposed benchmark in Table~\ref{exp:benchmark}, and the codes for reproducing the results are available at GitHub.

\section{Conclusion}
In this paper, we present AutoGL, the first library for automated machine learning on graphs, which is open-source, easy to use, and flexible to be extended. AutoGL is composed of a three-layer architecture with a fully automated pipeline covering five functional blocks, including auto feature engineering, neural architecture search, hyper-parameter optimization, model training, and auto ensemble. AutoGL is compatible with PyTorch Geometric and Deep Graph Library, two widely adopted graph machine learning libraries, and supports various tasks including node classification, link prediction, graph classification, heterogeneous node classification, etc. We also introduce AutoGL-light, a lightweight version of AutoGL by modularizing and focusing on core functions, which are more flexible and new user-friendly. NAS-Bench-Graph, the first dedicated benchmark for graph NAS, is also incorporated and introduced.

Currently, we are actively developing AutoGL. All kinds of inputs and suggestions are also warmly welcomed.

\ifCLASSOPTIONcompsoc
  \section*{Acknowledgments}
\else
  \section*{Acknowledgment}
\fi

This work was supported in part by the National Key Research and Development Program of China No. 2020AAA0106300, National Natural Science Foundation of China (No. 62250008, 62222209, 62102222, 62206149), Beijing National Research Center for Information Science and Technology under Grant No. BNR2023RC01003 and BNR2023TD03006, and Beijing Key Lab of Networked Multimedia. All opinions, findings, conclusions, and recommendations in this paper are those of the authors and do not necessarily reflect the views of the funding agencies.

\ifCLASSOPTIONcaptionsoff
  \newpage
\fi

\bibliographystyle{IEEEtran}
\bibliography{reference}

%

\begin{IEEEbiography}{Ziwei Zhang} received his Ph.D. from the Department of Computer Science and Technology, Tsinghua University, in 2021. He is currently a postdoc researcher in the Department of Computer Science and Technology at Tsinghua University. His research interests focus on machine learning on graphs, including graph neural networks (GNN), network embedding (a.k.a. network representation learning), and automated graph machine learning. He has published over 40 papers in prestigious conferences and journals, including KDD, NeurIPS, ICML, AAAI, IJCAI, and TKDE.
\end{IEEEbiography}

\begin{IEEEbiography}{Yijian Qin} received his B.E. from the Department of Computer Science and Technology, Tsinghua University in 2019. He is currently working toward a Ph.D. degree in the Department of Computer Science and Technology a Tsinghua University. His research interests include graph neural networks and neural architecture search. He has published several papers in prestigious conferences, e.g., ICML, NeurIPS, ICLR, ICDM, etc.
\end{IEEEbiography}

\begin{IEEEbiography}{Zeyang Zhang} received his B.E. from the Department of Computer Science and Technology, Tsinghua University in 2020. He is a Ph.D. candidate in the Department of Computer Science and Technology at Tsinghua University. His main research interests focus on graph representation learning, automated machine learning, and out-of-distribution generalization. He has published several papers in prestigious conferences, e.g., NeurIPS, AAAI, etc.
\end{IEEEbiography}

\begin{IEEEbiography}{Chaoyu Guan} obtained his master degree from the Department of Computer Science and Technology at Tsinghua University, supervised by Professor Wenwu Zhu. He got his B.E. degree in Computer Science and Technology from Shanghai Jiao Tong University. His research interests include graph representation learning, automated machine learning, neural architecture search, etc. He has published several high quality research papers in top-tier conferences including ICML, NAACL, CVPR workshop, ICLR workshop, etc. He has received the Tsinghua Top Grade Scholarship 2021 and Siebiel Scholarship 2021 in Tsinghua University.
\end{IEEEbiography}

\begin{IEEEbiography}{Jie Cai} received her master's degree from Tsinghua-Berkeley Shenzhen Institute at Tsinghua University in 2023, supervised by Professor Wenwu Zhu. She received her B.S. from the School of Mathematics, South China University of Technology in 2020. Her research interests focus on automated machine learning, graph learning, and social media. She has published 3 papers in AAAI and WWW.
\end{IEEEbiography}

\begin{IEEEbiography}{Heng Chang} received his Ph.D. Degree from the Tsinghua-Berkeley Shenzhen Institute at Tsinghua University in 2022. He received his B.S. from the Department of Electronic Engineering at Tsinghua University in 2017. His research interests focus on representation learning, adversarial robustness, and machine learning on graph/relational structured data. He has published several papers in prestigious conferences/journals including NeurIPS, AAAI, TheWebConf, TKDE, TPAMI, etc.
\end{IEEEbiography}

\begin{IEEEbiography}{Jiyan Jiang} received his Ph.D. in 2023 from the Department of Computer Science and Technology at Tsinghua University. His research interests include online learning and automated machine learning. He has published papers on relevant topics in top conferences including NeurIPS and ICLR.
\end{IEEEbiography}

\begin{IEEEbiography}{Haoyang Li} received his Ph.D. from the Department of Computer Science and Technology of Tsinghua University in 2023. Before that, he received his B.E. from the Department of Computer Science and Technology of Tsinghua University in 2018. His research interests are mainly in machine learning on graphs and out-of-distribution generalization. He has published high-quality papers in prestigious journals and conferences, e.g., TKDE, NeurIPS, KDD, IJCAI, ICLR, ACM Multimedia, etc.
\end{IEEEbiography}

\begin{IEEEbiography}{Zixin Sun} received his Master's degree in 2023 from the Tsinghua-Berkeley Shenzhen Institute at Tsinghua University. Currently, he works as a big data engineer in China Electronics Technology Group Corporation (CETC). His work primarily focuses on big data processing algorithms and systems.
\end{IEEEbiography}

\begin{IEEEbiography}{Beini Xie} received her M. Eng. in 2023 from the Tsinghua-Berkeley Shenzhen Institute at Tsinghua University. She received her B.S from the Statistical Department, Renmin University in 2020. Her research interests include adversarial robustness, machine learning, and neural architecture search on graph/relational structured data. She has published several papers in esteemed conferences and journals, including CVPR and TKDE. 
\end{IEEEbiography}

\begin{IEEEbiography}{Yang Yao} is currently a Ph.D. student at the Department of Computer Science and Technology at Tsinghua University. She got his B.E. degree from the Department of Computer Science and Technology at Tsinghua University in 2021. Her research interests include graph machine learning and automated machine learning.
\end{IEEEbiography}

\begin{IEEEbiography}{Yipeng Zhang} is currently a Ph.D. student at the Department of Computer Science and Technology at Tsinghua University. He got his B.E. degree from the Department of Computer Science and Technology at Tsinghua University in 2022. His research interests include machine learning, disentangled representation learning, and auxiliary learning.
\end{IEEEbiography}

\begin{IEEEbiography}{Xin Wang} is currently an Assistant Professor at the Department of Computer Science and Technology, Tsinghua University. He got both his Ph.D. and B.E degrees in Computer Science and Technology from Zhejiang University, China. He also holds a Ph.D. degree in Computing Science from Simon Fraser University, Canada. His research interests include multimedia intelligence and recommendations in social media. He has published over 100 high-quality research papers in top conferences and journals including ICML, NeurIPS, IEEE TPAMI, IEEE TKDE, ACM KDD, WWW, ACM SIGIR, ACM Multimedia, etc. He is the recipient of the 2020 ACM China Rising Star Award and the 2022 IEEE TCMC Rising Star Award.
\end{IEEEbiography}

\begin{IEEEbiography}{Wenwu Zhu} is currently a Professor in the Department of Computer Science and Technology at Tsinghua University, the Vice Dean of the National Research Center for Information Science and Technology, and the Vice Director of the Tsinghua Center for Big Data. Prior to his current post, he was a Senior Researcher and Research Manager at Microsoft Research Asia. He was the Chief Scientist and Director at Intel Research China from 2004 to 2008. He worked at Bell Labs New Jersey as a Member of Technical Staff during 1996-1999. He received his Ph.D. degree from New York University in 1996.
    
His current research interests are in the area of data-driven multimedia networking and Cross-media big data computing. He has published over 350 referred papers and is the inventor or co-inventor of over 50 patents. He received eight Best Paper Awards, including ACM Multimedia 2012 and IEEE Transactions on Circuits and Systems for Video Technology in 2001 and 2019.  
    
He served as EiC for IEEE Transactions on Multimedia (2017-2019). He served on the steering committee for IEEE Transactions on Multimedia (2015-2016) and IEEE Transactions on Mobile Computing (2007-2010), respectively. He serves as General Co-Chair for ACM Multimedia 2018 and ACM CIKM 2019, respectively. He is an IEEE Fellow, ACM Fellow, AAAS Fellow, SPIE Fellow, and a member of the Academy of Europe (Academia Europaea).
\end{IEEEbiography}

\end{document}